\documentclass{article}

\PassOptionsToPackage{numbers, compress}{natbib}
\usepackage[final]{neurips_2020_ml4ps}

\usepackage[utf8]{inputenc} % allow utf-8 input
\usepackage[T1]{fontenc}    % use 8-bit T1 fonts
\usepackage{hyperref}       % hyperlinks
\usepackage{url}            % simple URL typesetting
\usepackage{booktabs}       % professional-quality tables
\usepackage{amsfonts}       % blackboard math symbols
\usepackage{nicefrac}       % compact symbols for 1/2, etc.
\usepackage{microtype}      % microtypography
\usepackage{xcolor}
\usepackage{graphicx}
\usepackage{amsmath}
\usepackage{bbm}

\title{Instance Segmentation GNNs for One-Shot Conformal Tracking at the LHC}

\author{%
  Savannah Thais\\
  Princeton Institute for Computational Science and Engineering\\
  Princeton University\\
  Princeton, NJ 08540 \\
  \texttt{sthais@princeton.edu} \\
  \And
  Gage DeZoort\\
  Department of Physics \\
  Princeton University \\
  Princeton, NJ 08540 \\
  \texttt{jdezoort@princeton.edu} \\
}

\begin{document}

\maketitle

\begin{abstract}
3D instance segmentation remains a challenging problem in computer vision. Particle tracking at colliders like the LHC can be conceptualized as an instance segmentation task: beginning from a point cloud of hits in a particle detector, an algorithm must identify which hits belong to individual particle trajectories and extract track properties. Graph Neural Networks (GNNs) have shown promising performance on standard instance segmentation tasks. In this work we demonstrate the applicability of instance segmentation GNN architectures to particle tracking; moreover, we re-imagine the traditional Cartesian space approach to track-finding and instead work in a conformal geometry that allows the GNN to identify tracks and extract parameters in a single shot.  
\end{abstract}

\section{Introduction}
3D instance segmentation, the task of detecting object instances and their per-pixel segmentation mask, is a critical problem in computer vision. These input environments are often represented as point clouds. Traditional approaches to learning on point clouds relied on mapping the point cloud to a regular grid and using standard CNN methods, often resulting in sparsity or information loss. More recently, Graph Neural Networks (GNNs) have demonstrated excellent performance using unordered point sets as inputs \cite{deepsets}, and a few works have demonstrated the utility of GNNs for 3D object detection \cite{pointgnn}\cite{pointnet}. In this work, we propose a modified 3D instance segmentation GNN for the physics task of charged particle tracking at accelerators.

Recent work has shown that track edge classification via GNNs is a promising strategy for achieving scalable physics performance in both tracking and calorimetry \cite{exatrkx}. In these methods, a message-passing GNN predicts edge weights, which represent the probability that the edge belongs to a real particle trajectory. A secondary algorithm is used to build tracks from edge weights and extract track parameters. Our instance segmentation GNN architecture detects full tracks from a point cloud of hits and extracts track parameters in a single pass.

\subsection{Accelerated particle tracking}
The Large Hadron Collider (LHC) collides protons at energies of $13$ TeV, producing data crucial to precision measurements and searches for new physics. Particle physics detectors capture collision events using customized electronics systems that measure the interaction of the particles with the detector material. The readouts of these detectors are then algorithmically reconstructed into physics objects. In particular, it is necessary to reconstruct the trajectories (tracks) of individual particles from energy deposits in the tracker: the granular, multi-layered, inner-most part of the detector.

The tracker is embedded in a strong magnetic field that curves the trajectory of charged particles, resulting in helical tracks. These tracks can be uniquely defined by two parameters: the momentum in the transverse plane, $p_T$, and the transverse impact parameter $\epsilon_T$\footnote{This uniquely defines a track in the transverse plane, the transverse and longitudinal planes are independent.}. $p_T$ measures the particle's momentum in the plane perpendicular to the beamline, and is equal to $0.3\frac{GeV}{T*m}$BR where B is the strength of the magnetic field and R is the radius of the helix. $\epsilon_T$ measures the distance between the track's closest point of approach to the beamline and the beamline in the transverse plane; this is used to distinguish prompt particles (originating from the primary proton-proton collision) from displaced particles (originating from decays). 

\color{black}
\subsubsection{Conformal tracking}
In the standard conformal tracking scheme, circular particle trajectories satisfying $R^2=(x-a)^2 + (y-b)^2$ in the transverse plane are mapped onto $u$-$v$ space as follows: $u = \frac{x}{x^2+y^2}, \ \ v = \frac{y}{x^2+y^2}$
 Under this mapping, prompt tracks satisfying $R^2=a^2+b^2$ become straight lines of the form $v=\frac{1}{2b}-u\frac{a}{b}$. The authors in \cite{CMM} generalize this method to accommodate displaced tracks satisfying $\delta = R^2-a^2-b^2$. Assuming $\delta \ll R^2$, the $(x,y)\mapsto(u,v)$ map yields parabolas in $u$-$v$ space: $v = \frac{1}{2b} - u\frac{a}{b} - u^2\epsilon_T\big(\frac{R}{b}\big)^3$
 Accordingly, parabolic fits in $uv$-space directly yield transverse track parameters $a$, $b$, and $\epsilon_T$, where one approximates $R^2\approx a^2 + b^2$ to extract $\epsilon_T$. 

\section{Dataset}
To evaluate our architecture, we use the TrackML dataset \cite{TrackML}, a set of simplified high-luminosity particle collision events containing detector hits and ground truth information such as $p_T$ and particle ID ($p_{ID}$) about the particles that produced them. The TrackML detector is a generalized particle physics tracker, comprised of cylindrical detector layers immersed in a strong magnetic field. In particular, the pixel detector is a high-granularity sub-detector with a spatial resolution of $50\ \mu $m $\times\ 50\ \mu$m, which sits in the innermost region of the TrackML detector. To approximate the problem of track seeding in the planned high-pileup HL-LHC (high-luminosity LHC) conditions, we select only hits from the pixel detector.

\subsection{Graph construction}
\begin{figure}[htpb]
    \centering
    \includegraphics[width=0.32\textwidth]{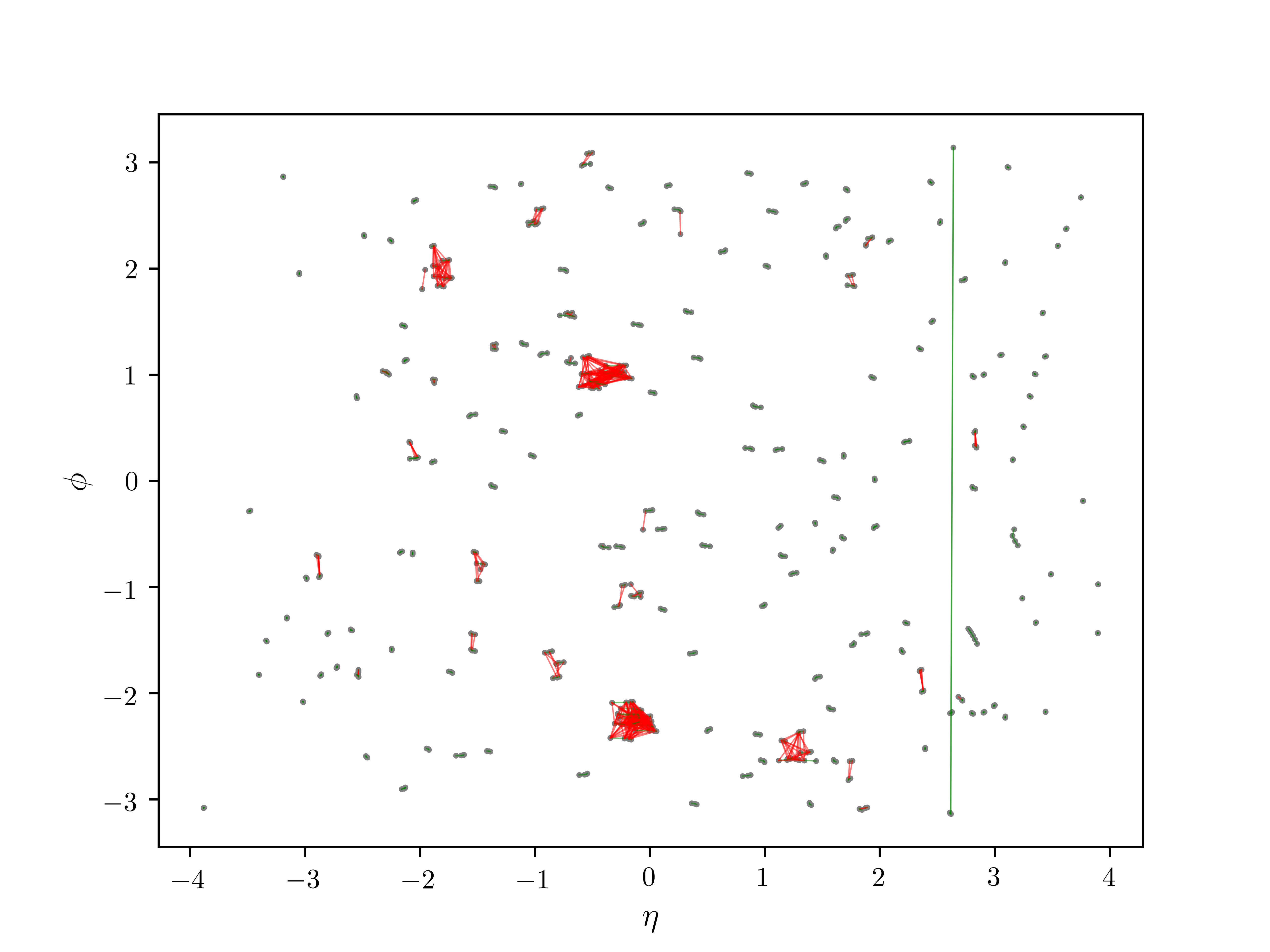}
    \includegraphics[width=0.32\textwidth]{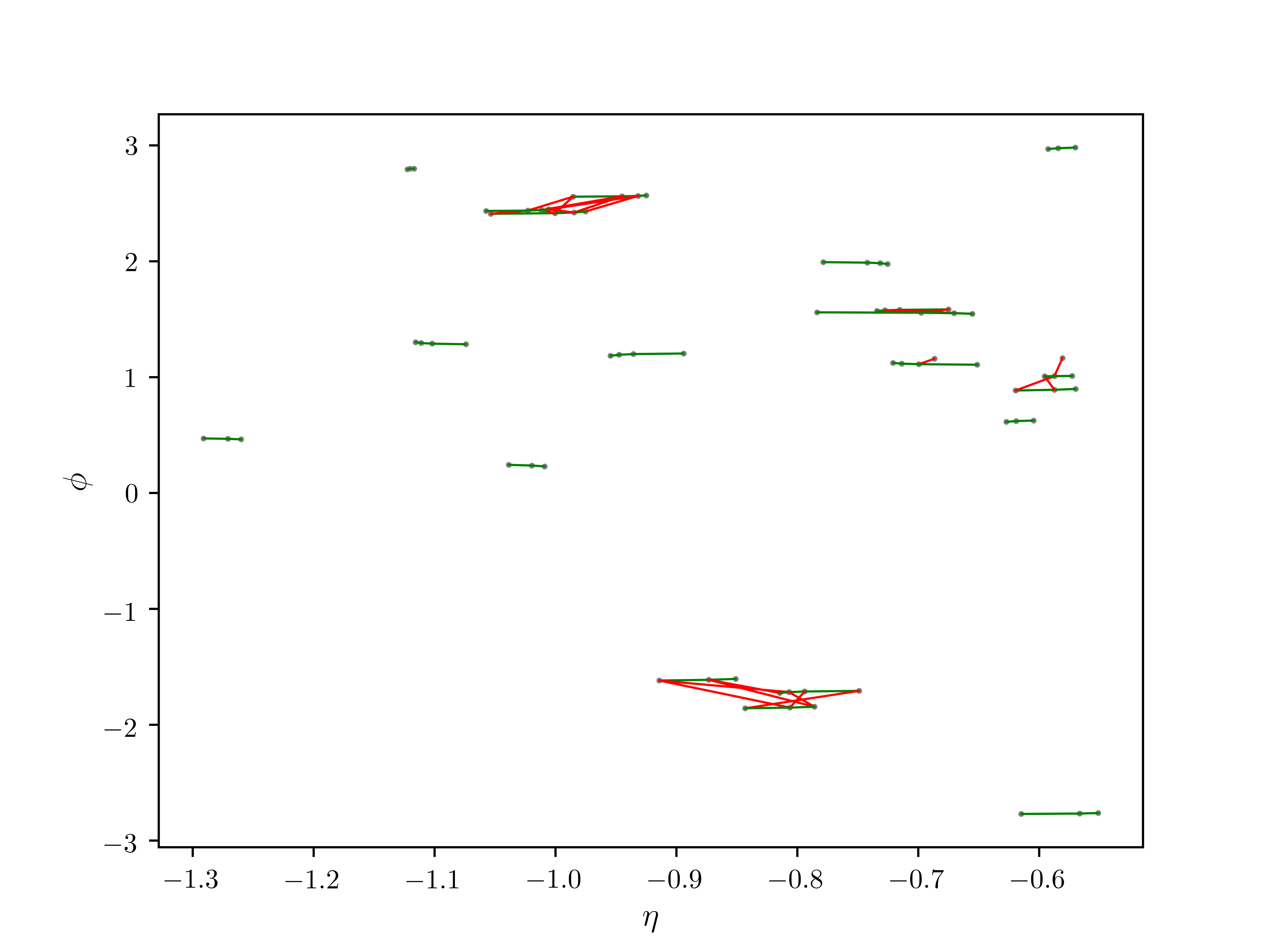}
    \includegraphics[width=0.32\textwidth]{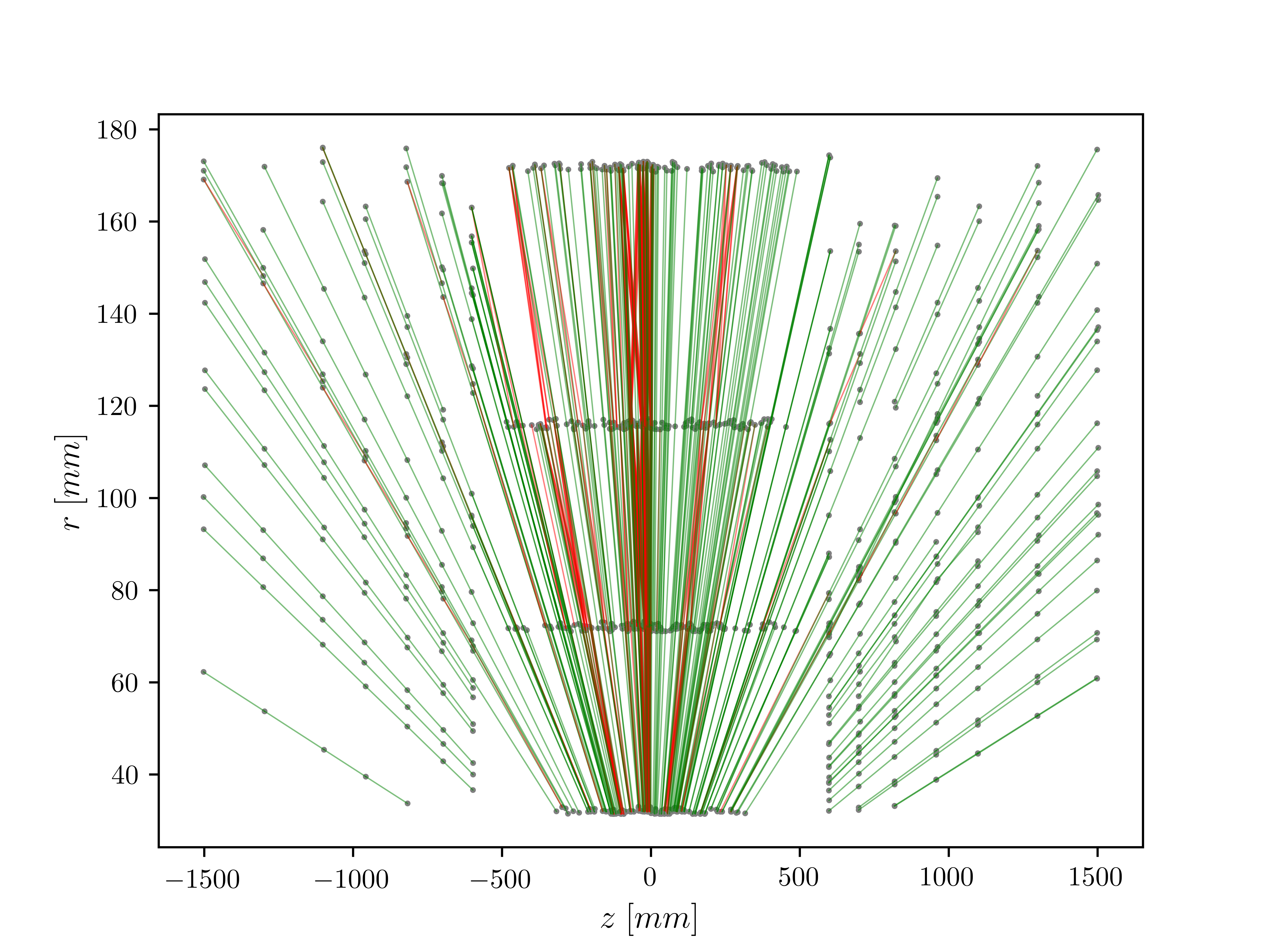}
    \caption{DBScan is used to cluster hits in $\eta$-$\phi$ space (left), yielding track-like graphs in $r$-$z$ space (right). True edges are colored green and false edges are colored red. Zooming into an $\eta$ window shows that many tracks are isolated, and the majority of false edges connect nearby tracks (center). }
    \label{fig:graphconstruction}
\end{figure}

%Maybe shorten this paragraph if we need space
Cleaned TrackML events are embedded as graphs by identifying hits as vertices and track segments, the lines connecting hits produced by the same particle, as edges. Our graph construction algorithm runs in three steps: track parametrization, track localization, and graph building. In the track parametrization step, particle truth trajectories in conformal space are fit with parabolas, generating a set of fit parameters $y\in\mathbb{R}^{n_{hits}\times3}$ for each hit. In the graph building step, the DBScan clustering algorithm \cite{DBScan} is used to cluster hits in $\eta$-$\phi$ space, where $\phi\in[0,2\pi)$ is the spherical azimuthal coordinate and $\eta\in(-\infty, \infty)$ a commonly used measure of the angle between a particle and the beamline, defined with respect to the spherical polar angle: $\eta = -\ln\big[\tan\big(\frac{\theta}{2}\big) \big]$. 
Crucially, particles traveling in similar directions are clustered in $\eta$-$\phi$ space (see Figure \ref{fig:graphconstruction}). The track localization stage is also run in $\eta-\phi$ space and generates a set of elliptical bounding boxes $y\in\mathbb{R}^{n_{tracks}\times5}$ around each truth track. 

The final graphs take the form $G=(P,E)$ with the point cloud $P=\{p_1,...,p_{n_{hits}}\}$, where $p_i=(\eta_i,\phi_i,s_i)$ is a hit with coordinates $\eta_i$ and $\phi_i$ and state value $s_i\in\mathbb{R}^2$, as the vertices, and $E=\{(p_i,p_j)\}$, the set of connections formed by DBScan, as the edges. We initialize the state value as $s_i=(z_i,l_i)$ where $z_i$ is the hit's original Cartesian space z coordinate and $l_i$ is the hit's layer number in the detector. 

\section{Conformal GNNs for one-shot tracking}
Our conformal GNN architecture is inspired by the Point-GNN architecture for bounding box-based instance segmentation \cite{pointgnn}. It contains three components: (a) a GNN of T iterations, (b) bounding-box localization, and (c) track parameter extraction. A diagram of the network is shown in Figure \ref{fig:diagrams}.

\subsection{Graph Neural Network with Auto-registration}
The GNN employs a standard message-passing process to update the state value of each vertex by aggregating features of its neighbors. In the $(t+1)^{th}$ iteration, each vertex state value is updated in the form:
\begin{equation}
\begin{gathered}
    s_i^{t+1}=g^t(\rho(\{f^t(x_j-x_i, s_j^t) \mid (i,j)\in E\}),s_i^t) \\    e_{ij}^t=f^t(v_i^t,v_j^t)
\end{gathered}
\end{equation}
where $e^t$ and $s^t$ are the edge features and state values from the $t^{th}$ iteration. The function $f^t()$ computes the edge features between a pair of vertices, $\rho$ aggregates the edge features of each vertex, and $g^t()$ updates the vertex's state value using the aggregated edge features. 

We also include an autoregistration mechanism to reduce the translation variance within neighborhoods. This mechanism aligns neighbors by their structural features: the vertex' state value from the previous iteration is used to predict and alignment offset $\Delta x_i^t=h^t(s_i^t)$ and Equation (1) becomes
\begin{equation}
    s_i^{t+1}=g^t(\rho(\{f(x_j-x_i+\Delta x_i^t, s_j^t)\}, s_i^t)
\end{equation}
In practice, the functions $f^t()$, $g^t()$, and $h^t()$ are approximated by multi-layer perceptrons ($MLP$) and $\rho$ is chosen to be $Max$. Each iteration $t$ is represented by a different set of $MLP^t$. After the $T^{th}$ iteration, the vertex state values are used to classify each vertex as a track hit or noise hit, compute an $\eta-\phi$ space elliptical bounding-box for each vertex, merge the bounding ellipses, and to predict the track parameters $(p_T,\epsilon_T)$. 

\subsection{Loss function}
The classification branch, $MLP_c$, computes a binary probability that the vertex belongs to a track or the background class. We use binary cross entropy as the classification loss: 
\begin{equation}
l_c=-\frac{1}{n_{hits}}\sum_{i=1}^{n_{hits}}y\log y_i+(1-y_i)\log(1-y_i)  
\end{equation}.

\begin{figure}[htpb]
    \centering
    \includegraphics[width=0.3\textwidth]{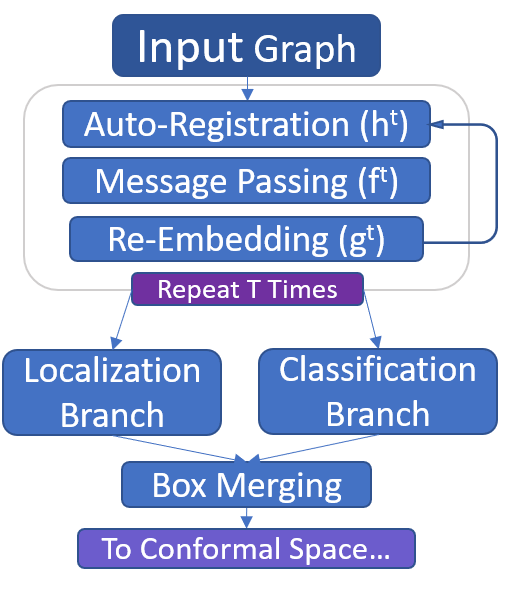}
    \hspace{1 cm}
    \includegraphics[width=0.4\textwidth]{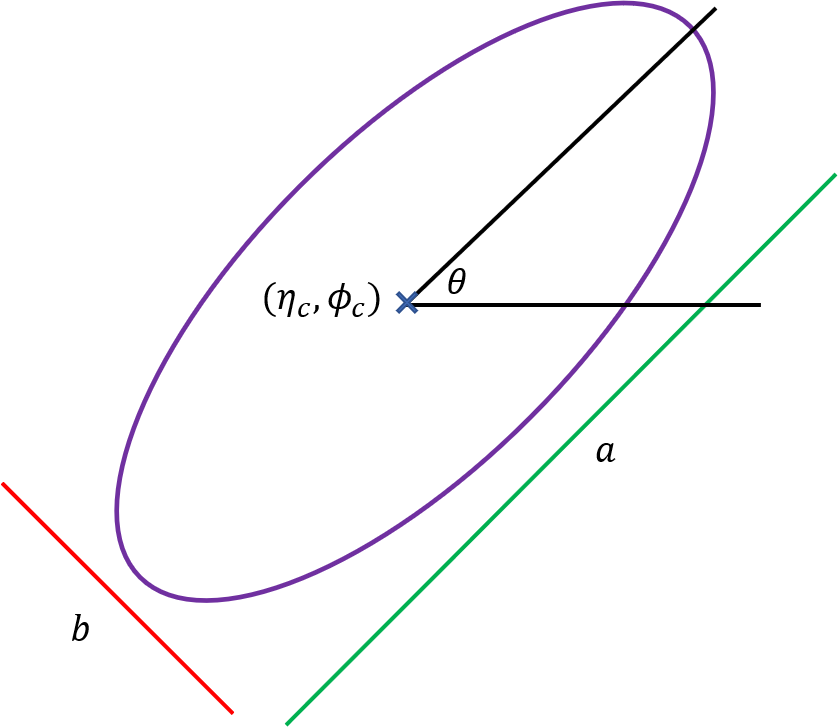}
    \caption{Left: Proposed GNN architecture. Right: 5 dof parametrization of elliptical bounding box.}
    \label{fig:diagrams}
\end{figure}

For the elliptical bounding boxes, we predict a 5 degree-of-freedom format $B=(\eta_c,\phi_c,a,b,\theta)$, where $(\eta_c,\phi_c)$ are the center of the ellipse, $(a,b)$ are its semi-major and semi-minor lengths, and $\theta$ is its rotation with respect to the $\eta$-axis (Figure \ref{fig:diagrams}). We encode the bounding-boxes with the vertex coordinates $(\eta_v,\phi_v)$ as 
\begin{equation}
\delta_{\eta}=\frac{(\eta_c-\eta_v)}{\eta_m}, \; 
\delta_{\phi}=\frac{(\phi_c-\phi_v)}{\phi_m}, \; 
\delta_{a}=log(\frac{a}{a_m}), \; 
\delta_{b}=log(\frac{b}{b_m}), \;
\delta_{\theta}=\frac{\theta+\Delta_{\theta}}{\theta_m}
\end{equation} 
where $\eta_m$, $\phi_m$, $a_m$, $b_m$, $\theta_m$, $\Delta_{\theta}$ are constant scale parameters. The localization branch, $MLP_{loc}$ predicts the encoded bounding ellipses for each vertex; if a vertex is classified as a track hit, we compute the Huber loss \cite{huber} between the true box and our prediction and we average over all the vertices: 
\begin{equation}
l_l=\frac{1}{n_{hits}}\sum_{i=1}^{n_{hits}} \mathbbm{1}(v_i\in\{trackhits\})l_{huber}(\delta - \delta^{gt}).    
\end{equation}.

The clusters formed by the classification and localization branches are transformed into conformal space and passed to the final tracking branch, $MLP_t$, which predicts the track parameters $(p_T,\epsilon_T)$. We use the standard MSE loss as the tracking loss:
\begin{equation}
l_t=\frac{1}{n_{clusters}}\sum_{i=1}^{n_{clusters}}(\frac{p_{T_i}-p_{T_i}^p}{c_{p_T}})^2+(\frac{\epsilon_{T_i}-\epsilon_{T_i}^p}{c_{\epsilon_T}})^2.    
\end{equation}

The final loss function is then $l_{total}=\alpha l_c + \beta l_{loc} + \gamma l_t$, where $\alpha$, $\beta$, and $\gamma$ are constant weights to balance each loss.

\section{Experiments}
The proposed architecture is implemented in PyTorch Geometric, a library extending PyTorch functionality to incorporate geometric deep learning \cite{PyG}. We use $T=4$ iterations, where each of the $g^t$ and $h^t$ MLPs has $1$ hidden layer with $64$ dimensions and a $ReLU$ activation, and an output of $2$ dimensions. The node embedding layer $f^t$ MLPs each have $2$ hidden layers with $ReLU$ activations, and an output of $4$ dimensions. The classifier branch has $3$ hidden layers with $ReLU$ activations, and a $2$ dimensional output with a $Sigmoid$ activation. The localization branch has $3$ hidden layers with $ReLU$ activations and a $4$ dimensional output. We set $(\eta_m,\phi_m,a_m,b_m,\theta_m,\Delta_{theta})=(0.01,0.004,0.038,0.005,\pi/4,0.5)$, values that were found to bring each ellipse parameter to the same scale. In each forward pass, $h^t$, $f^t$, and $g^t$ are called in sequence, with a residual connection between the node features. An Adam optimizer is used to facilitate loss optimization with a learning rate of $10^{-6}$ and weight decay of $10^{-5}$. The network is trained for $30$ epochs, at which point the network converges on a minimum. 

\begin{figure}[htpb]
    \centering
    \includegraphics[width=0.9\textwidth]{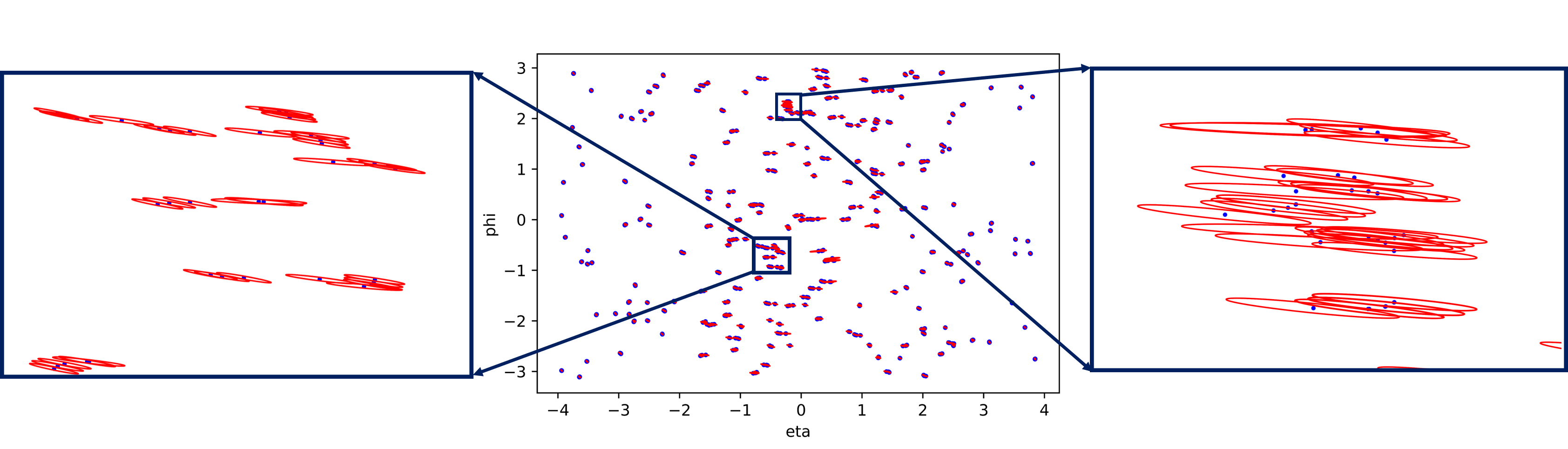}
    \caption{Center: a $p_T\geq$ 2GeV event in $\eta-\phi$ space with predicted ellipses for each graph node. Sides: zoomed in views of an eta/phi slice of the event}
    \label{fig:box_results}
\end{figure}

Initial results are shown in Figure \ref{fig:box_results} which shows that the predicted bounding ellipses effectively localize each track. Furthermore, the elongated nature of tracks in eta-phi space ensures that the ellipses overlap for same-track hits more frequently than for neighboring-track hits. Note that many ellipses have a similar orientation. We do not observe a uniform distribution of phi coordinates in the truth ellipses, which in turn biases the predicted ellipse orientations. Increasing model flexibility with respect to ellipse orientation is the subject of ongoing work.

\subsection{On-going work}
Our initial network achieves excellent track segmentation and demonstrates the feasibility and utility of this approach to tracking. However, this is a work in progress and we are exploring several methods to refine and improve our procedure and to demonstrate performance in more realistic pileup environments (e.g. by decreasing the $p_T$ threshold in graph construction). 

In particular, because tracks contain multiple hits, the GNN can predict multiple distinct bounding ellipses for a single track. It is necessary to merge these boxes into one and assign a confidence score. We are currently implementing a modification of the standard non-maximum suppression algorithm that averages over all overlapping bounding ellipses in a cluster whose Intersection-Over-The-Union exceeds some threshold $T_h$, where $T_h$ is optimized to separate overlapping ellipses from a single track from overlapping ellipses from distinct tracks. We can further extend this merging algorithm by incorporating information from the state values of the vertices.  

%We anticipate this will be more of an issue in high pile-up graphs and that standard techniques like non-maximum suppression will be insufficient. We are thus developing a custom merging and scoring algorithm that incorporates information from the state values of the vertices rather than just their classification score. 

Additionally, the architecture itself can likely be further optimized and we are systematically studying the size of the individual $MLP$s and relevant hyperparameters. In particular, the track parameter prediction network, $MLP_t$, has yet to be studied. 

%We will also consider other parameterizations of the bounding boxes (including flexible polygons rather that fixed rectangles). For higher density (lower $p_T$) graphs we can also include a Intersection Over Union term in the loss function to separate real and false hit clusters \cite{giou}. 

Finally, we hope to further exploit the conformal space transformation by re-defining the localization branch to predict bounding ellipses directly in conformal space, allowing the entire GNN architecture to operate in conformal space.

%directly predict parabolic fits in conformal space, rather than bounding boxes in $\eta-\phi$ space. The parabolic fit parameters can vary in orders of magnitude between tracks, and so preliminary studies of this method demonstrate a need for careful embedding and normalization of the parabola parameters to allow the GNN to effectively learn. Nonetheless, we believe this is a promising and extremely novel approach to particle tracking. 

\color{black}

\section{Conclusions}
We have re-framed the problem of particle tracking as an instance segmentation task and proposed a GNN architecture to detect tracks and extract their parameters in one shot. Our architecture exploits the $\eta-\phi$ and conformal symmetries inherent to particle physics detector data to simplify the learning task and reduce the computational load. Initial studies demonstrate promising performance and we are actively pursuing multiple methods of improvement.

\section*{Broader Impact}
This work may be used to improve the efficiency of particle tracking at the LHC and to enable the physics program of the High Luminosity LHC which requires redesigned software to process increased data loads. The physics results of the HL-LHC will yield many benefits to science and society. 

However, we are aware that this works exists within the broader context of computer vision, which has been used for many negative purposes including unlawful surveillance, military targeting, and biased policing. We are committed to continuing to learn about the broader use of computer vision and to support our communities in advocating against harmful uses. 

\begin{ack}
S.T. is supported by IRIS-HEP through the U.S. National Science Foundation (NSF) under Cooperative Agreement OAC-1836650. G.D. is supported by Department of Energy grant DE‐SC0007968.

We gratefully acknowledge the input and discussion from the ExaTrkX collaboration, as well as Isobel Ojalvo and Lindsey Gray. \\ \\ \\ \\ \\
\end{ack}

%\begin{thebibliography}{9}
%\bibitem{CMM} 
%\end{thebibliography}

\bibliographystyle{lucas_unsrt}
\bibliography{references}

\end{document}